 \documentclass[pmlr,twocolumn,10pt]{jmlr} 

\usepackage{booktabs}
\usepackage{siunitx}

\usepackage[switch]{lineno}
\usepackage{listings}
\usepackage{xcolor}
\usepackage{multirow}

\definecolor{codegreen}{rgb}{0,0.6,0}
\definecolor{codegray}{rgb}{0.5,0.5,0.5}
\definecolor{codepurple}{rgb}{0.58,0,0.82}
\definecolor{backcolour}{rgb}{0.95,0.95,0.95}

\lstdefinestyle{mypython}{
    language=Python,
    backgroundcolor=\color{backcolour},
    commentstyle=\color{codegreen}\ttfamily,
    keywordstyle=\color{blue}\bfseries,
    numberstyle=\tiny\color{codegray},
    stringstyle=\color{codepurple},
    basicstyle=\ttfamily\footnotesize,
    breaklines=true,
    numbers=left,
    numbersep=5pt,
    showstringspaces=false,
    tabsize=4
}
\theorembodyfont{\upshape}
\theoremheaderfont{\scshape}
\theorempostheader{:}
\theoremsep{\newline}

\jmlrvolume{297}
\jmlryear{2025}
\jmlrworkshop{Machine Learning for Health (ML4H) 2025} 

 \title[cSCC Grading using PathFMTools]{Leveraging Foundation Models for Histological Grading in Cutaneous Squamous Cell Carcinoma using PathFMTools}

\author{%
\Name{Abdul Rahman Diab}\Email{abdul@ds.dfci.harvard.edu}\\
\addr Dana-Farber Cancer Institute\\
\AND
\Name{Emily E. Karn}\Email{emilyekarn@gmail.com}\\
\addr Brigham and Women's Hospital\\
\AND
\Name{Renchin Wu}\Email{renchin\_wu@dfci.harvard.edu}\\
\addr Dana-Farber Cancer Institute\\
\AND
\Name{Emily S. Ruiz}\Email{esruiz@bwh.harvard.edu}\\
\addr Brigham and Women's Hospital\\
\AND
\Name{William Lotter} \Email{lotterb@ds.dfci.harvard.edu}\\
\addr Dana-Farber Cancer Institute, Brigham and Women's Hospital, \& Harvard Medical School 
}

\begin{document}

\maketitle

\begin{abstract}
Despite the promise of computational pathology foundation models, adapting them to specific clinical tasks remains challenging due to the complexity of whole-slide image (WSI) processing, the opacity of learned features, and the wide range of potential adaptation strategies. To address these challenges, we introduce \textbf{PathFMTools}, a lightweight, extensible Python package that enables efficient execution, analysis, and visualization of pathology foundation models. We use this tool to interface with and evaluate two state-of-the-art vision-language foundation models, CONCH and MUSK, on the task of histological grading in cutaneous squamous cell carcinoma (cSCC), a critical criterion that informs cSCC staging and patient management. Using a cohort of 440 cSCC H\&E WSIs, we benchmark multiple adaptation strategies, demonstrating trade-offs across prediction approaches and validating the potential of using foundation model embeddings to train small specialist models. These findings underscore the promise of pathology foundation models for real-world clinical applications, with \textbf{PathFMTools} enabling efficient analysis and validation.
\end{abstract}

\paragraph*{Data and Code Availability}
The cSCC dataset utilized has not been IRB-approved for public release. The PathFMTools code is available at \url{https://github.com/lotterlab/pathfmtools}. 
\paragraph*{Institutional Review Board (IRB)}
This research was approved by the Mass General Brigham IRB (protocol 2023P002386)  

\section{Introduction}
\label{sec:intro}

Foundation models have emerged as powerful tools in computational pathology, offering the ability to extract rich, domain-specific features from histopathology slides~\citep{li2025review, bilal2025review}. By training on large collections of digitized H\&E slides with self-supervised learning, these models learn generalizable morphological patterns that can be leveraged for downstream applications even when data is scarce. Nevertheless, adapting foundation models to specific research or clinical contexts presents several  challenges. First, their use often involves complex, disjointed pipelines spanning data pre-processing, model inference, and downstream analysis of hundreds or thousands of gigapixel-scale whole-slide images (WSIs). Second, the feature spaces learned by foundation models remain opaque, necessitating efficient probing tools to facilitate biological validation. Third, the wide array of potential adaptation strategies --- ranging from zero-shot inference to supervised transfer learning --- underscores the importance of benchmarking studies that compare these approaches in task-specific settings.

One particularly important task for many types of cancer is histological grading. Grading systems assess how differentiated tumor cells appear relative to normal cells --- a key indicator of tumor aggressiveness and clinical risk. In cutaneous squamous cell carcinoma (cSCC), a highly prevalent skin cancer with over one million diagnoses annually in the United States, tumor grade is a critical component of risk stratification~\citep{Que2018-yq, Thompson2016-oo}. Although most cSCC tumors are successfully managed with local excision, a subset of patients experience metastasis and/or disease-specific death~\citep{Schmults2013-qi}. Notably, a recent large, multi-institutional study found that among several existing risk factors, tumor grade was the strongest predictor of patient outcomes~\citep{Jambusaria-Pahlajani2025-vh}. Nevertheless, grading remains a subjective and error-prone process, exhibiting limited reproducibility and high inter- and intra-rater variability~\citep{Nash2022-dd}. Based on this, the American Joint Committee on Cancer has excluded differentiation from the 8th edition staging for cSCC despite its prognostic significance, though it is included in the popular BWH staging system. Thus, the clinical significance of cSCC grading and the associated challenges highlight an important opportunity to assess the utility of foundation models.

In this work, we introduce \textbf{PathFMTools}, a lightweight Python package that enables efficient exploration and analysis of computational pathology foundation models. We leverage the package to evaluate two state-of-the-art vision-language foundation models, CONCH~\citep{lu2024avisionlanguage} and MUSK~\citep{Xiang2025-ux}, on the task of cSCC grading. Using a curated cohort of 440 cSCC WSIs, we benchmark multiple adaptation strategies and explore the biological relevance of the learned features. 

\section{Related Work}
\label{sec:relw}

\subsection{Foundation Models in Computational Pathology}
Numerous foundation models have been developed for computational pathology, and the available options differ in their training datasets, self-supervised learning strategies, and model outputs~\citep{li2025review, bilal2025review}. The most common class of models operates at the patch level, where fixed-size crops (e.g. $224 \times 224$ pixels) from H\&E-stained WSIs are used as inputs to generate feature embedding vectors. These embeddings can subsequently be used for a variety of downstream tasks, including slide-level prediction via aggregation across all patches.

Two predominant self-supervised learning paradigms have emerged for training such models: (1) vision-language contrastive learning and (2) vision-only self-distillation with masked data augmentation. Vision-language contrastive learning uses paired image and text inputs to learn visual and textual representations in a shared space~\citep{clip}. This approach enables zero-shot classification, where image embeddings are compared to text embeddings to perform inference without task-specific training. In contrast, popular self-distillation methods such as DINOv2~\citep{dinov2} rely solely on image data, encouraging the model to learn stable representations under data augmentation and masking.

Given the zero-shot capabilities of vision-language models, we focus our analysis on two publicly-available pathology vision-language foundation models: CONCH and MUSK. CONCH was trained on approximately 1.17 million image-caption pairs sourced from both public and private datasets~\citep{lu2024avisionlanguage}, and MUSK was trained on over 50 million pathology images and over one billion pathology-related text tokens from publicly-available sources~\citep{Xiang2025-ux}. Although our analyses are centered on these two models, the software package we introduce is model-agnostic and compatible with a broad range of foundation model architectures.

\subsection{Software Packages for Computational Pathology}
Growing interest in computational pathology has led to the development of several open-source software packages in the field. Notable examples include TIAToolbox~\citep{Pocock2022}, STAMP~\citep{ElNahhas2024}, TRIDENT~\citep{zhang2025standardizing}, and PathML~\citep{pathml}, which vary according to functionality and design. Some earlier packages (e.g., TIAToolbox) support tasks such as cell and tissue segmentation but lack native compatibility with pathology foundation models. More recent tools (e.g., TRIDENT) enable foundation model embedding generation but place less emphasis on downstream analysis and visualization. PathFMTools offers both built-in embedding generation and exploratory analysis tailored to pathology foundation models, including unique text embedding and zero-shot classification modules for vision–language models.

\subsection{cSCC Grading}
Although no standardized grading system for cSCC currently exists~\citep{Nash2022-dd}, tumors are typically categorized as well, moderately, or poorly differentiated based on their histological appearance. In a recent multinational cohort study comprising over 20,000 tumors, histological grade emerged as the most prognostic factor for poor clinical outcomes~\citep{Jambusaria-Pahlajani2025-vh}, underscoring its clinical relevance in risk stratification.

Despite its importance, cSCC grading is highly subjective. Reported inter-rater agreement remains moderate, with Cohen's kappa values in the range of $\kappa \sim 0.53\text{-}0.55$ ~\citep{Nash2022-dd}. This variability highlights the need for more objective, quantitative grading approaches. Recent work has begun to explore machine-learning based solutions to this problem. For example, \cite{Choudhary2023-rt} developed and tested a weakly-supervised model trained from scratch to estimate tumor grade using 718 H\&E-stained WSIs from a single clinical site.

\section{Methods}

\subsection{The PathFMTools Package}
Here, we present \textbf{PathFMTools}, a lightweight, extensible package for running pathology foundation models and interfacing with the embeddings that they produce. 

\begin{figure*}[h]
\centering 
\includegraphics[width=0.8\textwidth]{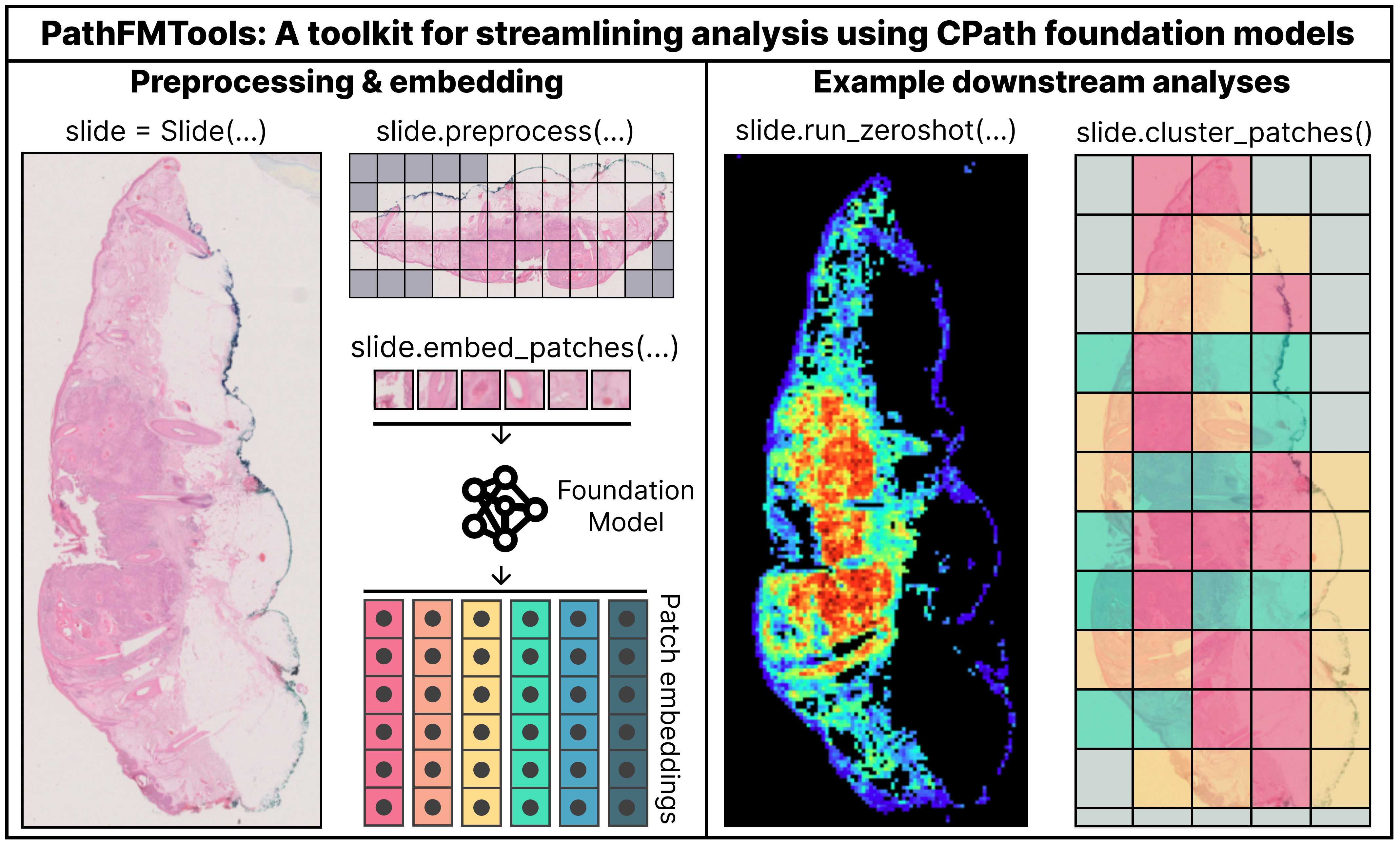} 
\vspace{-10pt}
\caption{Overview of core interactions supported by the \textbf{PathFMTools} package. The package implements a modular end-to-end pipeline that facilitates embedding generation and exploratory analysis, including zero-shot similarity visualization and clustering of embeddings. See \ref{apd:Code} for code examples.}
\label{fig:package_overview}
\end{figure*}

\subsubsection{Command line tools for efficient embedding generation}
The \textbf{PathFMTools} command line interface provides a simple and scalable entry point for executing core computational pathology workflows on collections of whole-slide images, enabling users to preprocess slides and generate foundation model embeddings using a single command.

To support large-scale data processing, the pipeline includes built-in support for automatic parallelization to an arbitrary number of workers. This scalability is enabled by the package's low memory footprint, achieved through efficient chunked processing of WSIs. Additionally, the workflow can be separated into distinct stages: a CPU-intensive preprocessing step and a GPU-dependent embedding generation step, facilitating the efficient use of computational resources in environments with limited or shared GPU availability.

\subsubsection{Modular Python design}
The command line interface is built upon the Slide class, which is the primary component that users working within Python will use. This class provides a unified abstraction for interacting with individual WSIs, encapsulating both the raw image data and any intermediate representations and outputs produced during analysis.

The class supports core computational pathology operations, such as tissue segmentation and tiling (Figure \ref{fig:package_overview}). By default, Otsu segmentation~\citep{otsu} is utilized, but the modular design enables custom methods. The segmentation and tiling operations are optimized to run on localized regions of the slide, enabling memory-efficient processing of gigapixel-scale images on standard consumer hardware.

\subsubsection{Built-in analysis and visualization}
In addition to preprocessing, the Slide class interfaces directly with foundation models through an encapsulated model class. This design enables patch-level embedding generation and zero-shot analysis with vision–language prompts (Figure \ref{fig:package_overview}), supporting downstream tasks such as clustering and multiple-instance learning. The package also implements native K-means clustering to aggregate embeddings across slides, producing both cluster centroids and discrete patch assignments.

The Slide class further provides visualization tools for exploratory analysis, including high- or low-resolution slide rendering, contextual tile inspection, random sampling, and heatmaps of zero-shot scores that map image–text similarity. These utilities support both quantitative analysis and qualitative interpretation of model behavior.

In addition to the Slide class, \textbf{PathFMTools} also provides the SlideDataset class, a custom subclass of the Pytorch Dataset object. This class enables the efficient integration of Slide objects into deep learning pipelines. It supports both pixel-based and embedding-based inputs and handles memory management and indexing internally, streamlining model training workflows.

\subsection{cSCC Analysis}
To evaluate how pathology foundation models might support real-world clinical workflows, we conducted a series of analyses reflecting common tasks that clinicians and researchers might wish to perform using model-generated embeddings. These tasks include zero-shot classification based on textual descriptors, unsupervised clustering for morphologic feature discovery, and weakly supervised classification using slide-level labels. Each approach is designed both to probe the different ways that the user can interact with foundation models --- ranging from minimal supervision to deep learning model training --- and to assess the utility of such models for histological grading of cSCC slides. To evaluate each approach, we formulated a binary classification task according to whether each slide corresponded to a well-differentiated (`0') or moderately/poorly differentiated (`1') tumor, though we also consider different thresholds in Appendix \ref{apd:cscc_grading}.

\begin{figure*}[h]
\centering 
\includegraphics[width=\textwidth]{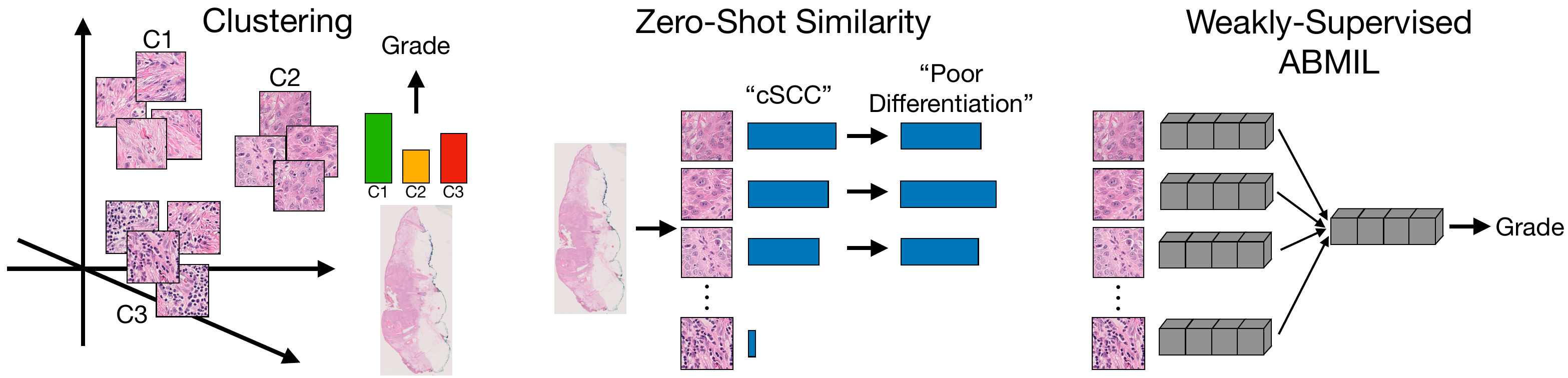} 
\caption{Summary of analytical approaches. Three different approaches were used to analyze the foundation model embeddings and their ability to assist with cSCC grading. Left: K-means clustering in the embedding space. Middle: A zero-shot classification task was formulated by first classifying each patch in a WSI as `cSCC' or `non-neoplastic' based on similarity between the patch embedding and text embeddings of these terms. Patches classified as cSCC were subsequently assessed for their similarity with text embeddings for well and poor differentiation. Right: Attention-based multiple instance learning (ABMIL) based on the patch embeddings was employed for a weakly-supervised benchmark.}
\label{fig:methods_overview} 
\end{figure*}

\subsubsection{Patch \& text embedding}
Each WSI in the dataset was tiled into non-overlapping $(448 \times 448)$ pixel patches at $40 \times$ magnification. To exclude background regions, we applied Otsu thresholding~\citep{otsu} to the distribution of mean grayscale pixel intensities across the slide's patches. Patches passing the tissue threshold were then downsampled to $(224 \times 224)$ pixels, the expected size of the studied foundation models, and passed through the respective model-specific preprocessing pipelines.

We generated patch-level embeddings using two publicly-available foundation models: CONCH and MUSK. Each model can produce two types of embeddings per patch: one optimized for downstream machine learning tasks (e.g. classification, regression), and one optimized for zero-shot tasks via alignment with pretrained language encoders. We use the machine learning-optimized embeddings for all analyses detailed in this work, except for the zero-shot approach. We generated text embeddings using the respective language encoders and tokenization pipelines provided by each foundation model.

\subsubsection{Zero-shot histological grading}
To assess the zero-shot capabilities of the foundation models in estimating WSI-level grading, we formulated a two-stage inference procedure using the language-aligned embeddings from CONCH and MUSK. In the first stage, we identified candidate cancer regions within each WSI by computing the cosine similarity between each patch embedding and the language embeddings of the phrases ``cutaneous squamous cell carcinoma" and ``non-neoplastic". Each patch was thus assigned two similarity scores (logits), and we retained only those patches for which the similarity to ``cutaneous squamous cell carcinoma" exceeded that to ``non-neoplastic", indicating that the patch was more likely to be neoplastic according to the model. Formally, for a slide $S_i$ with $m$ patches $S_i = \{p_1^{(i)},\ \dots,\ p_m^{(i)}\}$, we define the set of cancer-predicted patches as $$C_i := \{p_j^{(i)} \in S_i\ :\ P(\text{cancer}\ |\ p_j^{(i)}) > 0.5\}$$ 
In the second stage, we computed a grading score for each of retained cancer patches $C_i$ that reflects its differentiation on a continuum from well-differentiated to poorly-differentiated. We first computed the cosine similarity between each patch embedding and the language embeddings for the phrases ``poorly differentiated" and ``well differentiated". The grading score was then formulated as the difference in the two similarity scores for each patch. Formally given a patch $k$ with embedding $z_k^{(i)}$ and text embeddings $z_{poor}$ and $z_{well}$, the patch-level score $s_k^{(i)}$ is defined as $$s_k^{(i)} := \cos(z_k^{(i)},\ z_{poor}) - \cos(z_k^{(i)},\ z_{well})$$
Intuitively, a high value for $s_k^{(i)}$ indicates that the pixel embedding for patch $k$ is more aligned with features associated with poor differentiation.
Finally, we aggregate the patch-level scores within each slide into a slide-level score $s_i$ by taking the maximum over the cancer patch grading scores $$s_i := \max_{k \in C_i}\ s_k^{(i)}$$
This max-pooling approach reflects the clinical intuition that tumor grade is often determined by the most poorly-differentiated region within the tissue. We also evaluated alternative approaches for computing slide-level scores, including varied patch-level aggregation methods and prompt ensembling, which we present in Appendix \ref{apd:prompt}.

\subsubsection{K-means clustering of patch embeddings}
We applied K-means clustering with 25 clusters, selected based on silhouette analysis, to all patch embeddings pooled across the dataset, assigning each patch to one of 25 discrete groups. Each slide was then represented by a 25-dimensional normalized histogram, where the $j^{th}$ component quantifies the proportion of patches from that slide assigned to cluster $j$. This cluster histogram encodes the distribution of morphologic patterns captured by the foundation model across each slide. To evaluate the predictive utility of these representations, we applied two strategies:
\begin{itemize}
    \item Univariate AUROC: For each component of the 25-D summary vectors, we compute the area under the receiving operator characteristic curve (AUROC) for predicting the binarized histological grade. This provides an estimate of the standalone predictive power of individual clusters.
    \item Multivariate logistic regression: We train an $\ell_2$-regularized logistic regression model using the full 25-dimensional cluster histogram as input. This evaluates the collective predictive power of the cluster distribution.
\end{itemize}
    
\subsubsection{Attention-based multiple instance learning}
To further assess the utility of patch-level embeddings for predicting histological grade, we implemented an attention-based deep multiple instance learning (ABMIL) framework~\citep{Ilse2018-ph}. In this setting, each WSI is treated as a bag of instances (patch embeddings), and the model is trained to aggregate information across patches to produce a slide-level prediction without requiring patch-level labels.

We used a gated single-headed attention mechanism to compute a weighted aggregation of patch embeddings into a fixed-dimensional slide-level representation (see \ref{apd:abmil_math} for details).
This aggregated embedding is passed to a lightweight multilayer perceptron consisting of two hidden layers with ReLU activations ($64$ and $32$ neurons, respectively), followed by a sigmoid-activated scalar output node that predicts the tumor differentiation. We train the ABMIL model end-to-end using binary cross-entropy loss on the binarized grade labels (see \ref{apd:abmil_hyper_cross} for training details). We also visualize the attention scores assigned to individual patches.

\subsubsection{Cohort}

We curated the data used in this study from a U.S. academic hospital system, sampling 440 cSCC tumors from the system's tumor registry and scanning the associated H\&E slides for this work. Given the cost of retrieval and digitization, our data collection procedure consisted of retrieving all moderately-differentiated and poorly-differentiated tumors available in the registry, in addition to a random subset of the well-differentiated tumors, as high-grade tumors are comparatively rare and valuable for model evaluation and development. Differentiation level was determined based on the associated pathology report, and if not explicitly mentioned in the report, well-differentiated was assumed based on clinical convention~\citep{Jambusaria-Pahlajani2025-vh}. The resulting dataset contained 239 well-differentiated tumors, 99 moderately-differentiated tumors, and 102 poorly-differentiated tumors  (Table \ref{cohort_counts}). The majority of slides corresponded to diagnostic biopsies, with some slides corresponding to excisional biopsies. Only one slide was scanned per sampled tumor when multiple slides were available, which was selected randomly from the available slides. The WSIs were digitized at 40x magnification using a Hamamastu NanoZoomer digital slide scanner. See \ref{apd:cohort} for a detailed summary of the cohort, including patient and tumor characteristics.

\subsubsection{Statistical analysis}

AUROC was used as the primary assessment metric. Sensitivity and specificity at selected operating points for the ABMIL models are also presented in Appendix \ref{apd:cscc_grading}. For all analyses other than zero-shot performance quantification, we conducted five-fold cross-validation at the patient level and report performance as the average across folds (see \ref{apd:abmil_hyper_cross} for additional details). For the logistic regression models based on K-means cluster proportions, both the clustering and regression were fit per each fold before evaluation on the test split for the fold.

\section{Results} 
We evaluated the utility of foundation model embeddings for cSCC grading using three approaches: zero-shot classification, unsupervised clustering with K-means, and ABMIL. 

\subsection{Zero-shot histological grading}
We leveraged the language-aligned embeddings from CONCH and MUSK to assess zero-shot grading performance without task-specific fine-tuning. Using phrase-based similarity scores aggregated at the slide level (see Methods), this approach achieved AUROC values of $0.63$ and $0.60$ (Figure \ref{fig:auroc_logistic_abmil}), for CONCH and MUSK respectively, indicating performance moderately above chance.

\subsection{K-means clustering of patch embeddings}
\begin{figure}
    \centering
    \includegraphics[width=1\linewidth]{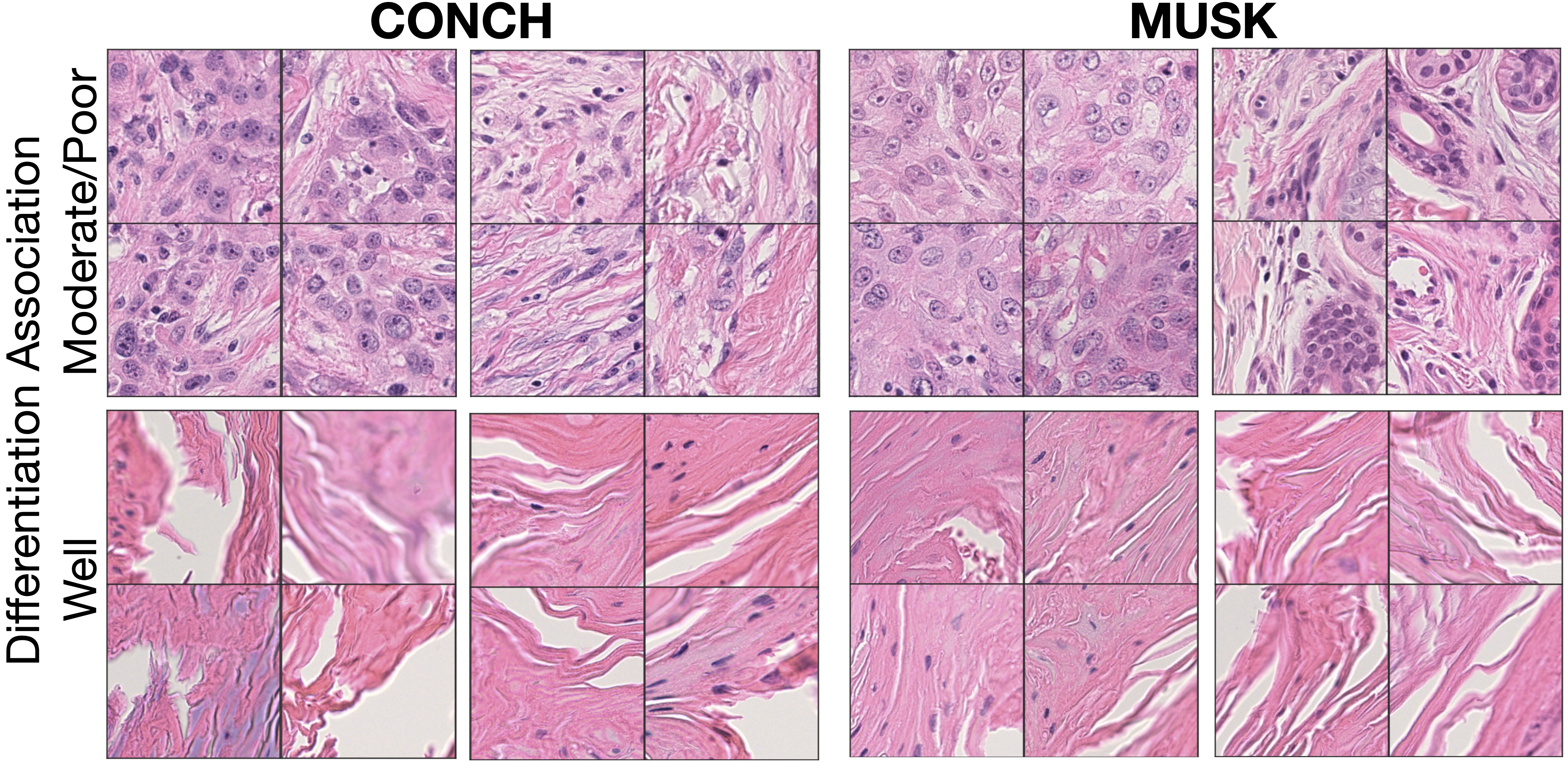}
    \vspace{-10pt}
    \caption{Examples of clusters most associated with histological grade for CONCH and MUSK. For each cluster, the patches whose embeddings are most similar to the cluster centroid are shown.}
    \label{fig:cluster_examples}
\end{figure}

Unsupervised clustering provided an alternative, interpretable representation of slide features by discretizing patches into clusters that capture general morphological patterns (Figure \ref{fig:cluster_examples}, \ref{fig:conch_clusters_full}, \ref{fig:musk_clusters_full}). 
By clustering all patch embeddings into 25 clusters, we obtained a normalized histogram of cluster proportions for each slide. We then computed univariate AUROCs using individual histogram components as scores and multivariate AUROC by training logistic regression models on the histograms. 

Univariate AUROC analysis demonstrated that individual clusters can be surprisingly informative, achieving AUROCs of $\sim$0.75 for the most-discriminative clusters. Inspection of these clusters yielded mostly intuitive findings (Figure \ref{fig:cluster_examples}). For associations with the high-grade label, both CONCH and MUSK exhibited clusters of moderately/poorly differentiated cancer cells. CONCH also exhibited a strongly-discriminative cluster corresponding to desmoplasia, a known risk factor potentially linked to higher grade~\citep{Que2018-yq}. However, one of the MUSK clusters most associated with high-grade cSCC corresponded to sweat glands, an association that is unclear. Clusters associated with low-grade tumors frequently contained keratin fragments, consistent with the keratinization typical of well-differentiated cSCC. 

\begin{figure}[h]
    \centering
    \includegraphics[width=1.0\linewidth]{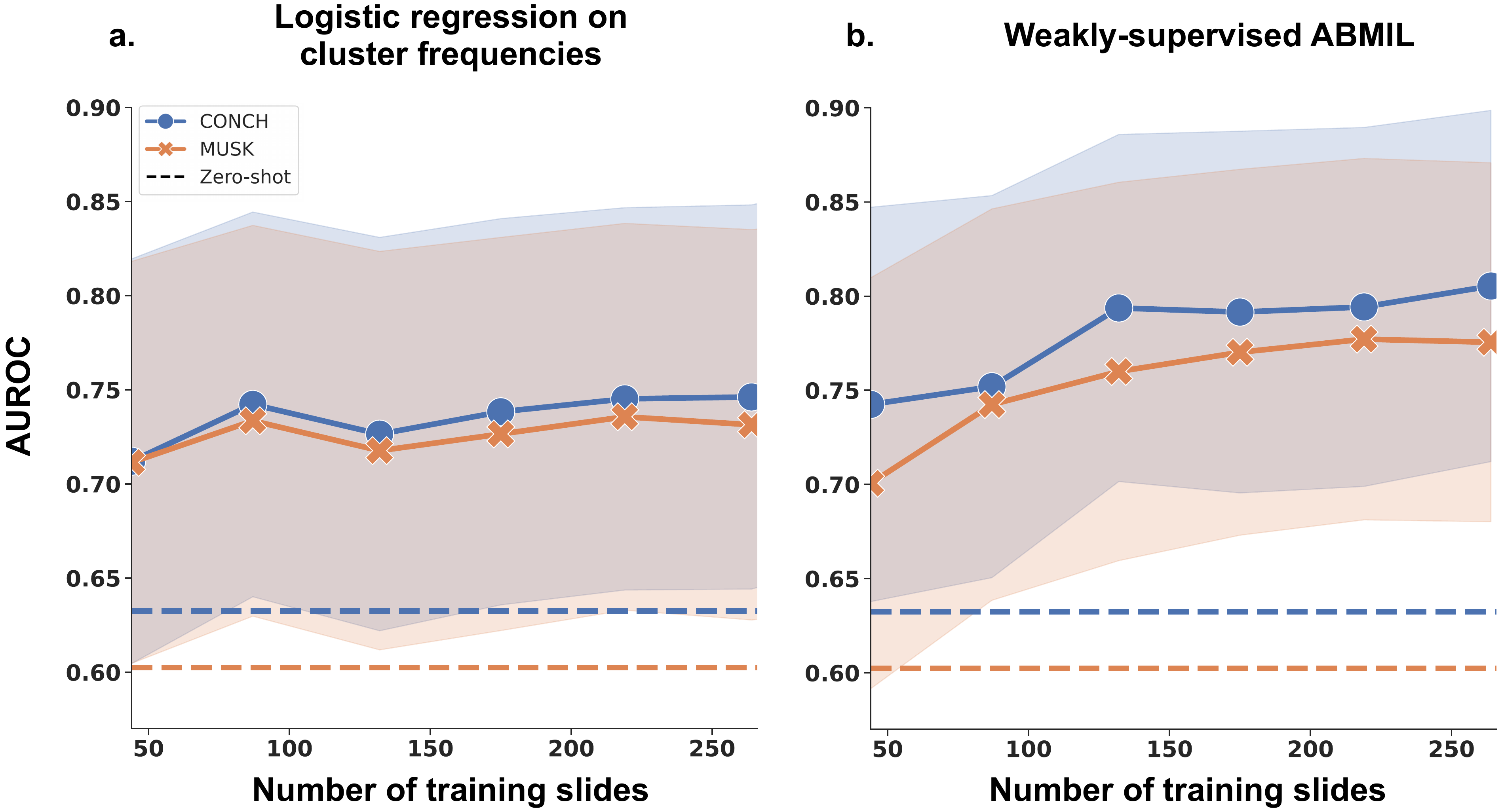}
    \caption{Grade classification performance. The dashed lines represent the zero-shot classification performance leveraging the vision-language capabilities of CONCH and MUSK. The markers represent the mean test AUROC of models trained using CONCH and MUSK patch embeddings across the 5 test splits for each number of training slides. Error bars represent DeLong 95\% confidence intervals. \textbf{(a)} Performance of logistic regression trained on the cluster frequency vectors. \textbf{(b)} Performance of ABMIL models.}
    \label{fig:auroc_logistic_abmil}
\end{figure}

Multivariate logistic regression on all 25 cluster frequencies did not substantially exceed the performance of the best univariate clusters (Figure \ref{fig:auroc_logistic_abmil}), though it outperformed the zero-shot strategy and was assessed in a stricter cross-validation setting.  


\subsection{Attention-based multiple instance learning (ABMIL)}
\begin{figure}[h]
    \centering
    \includegraphics[width=1\linewidth]{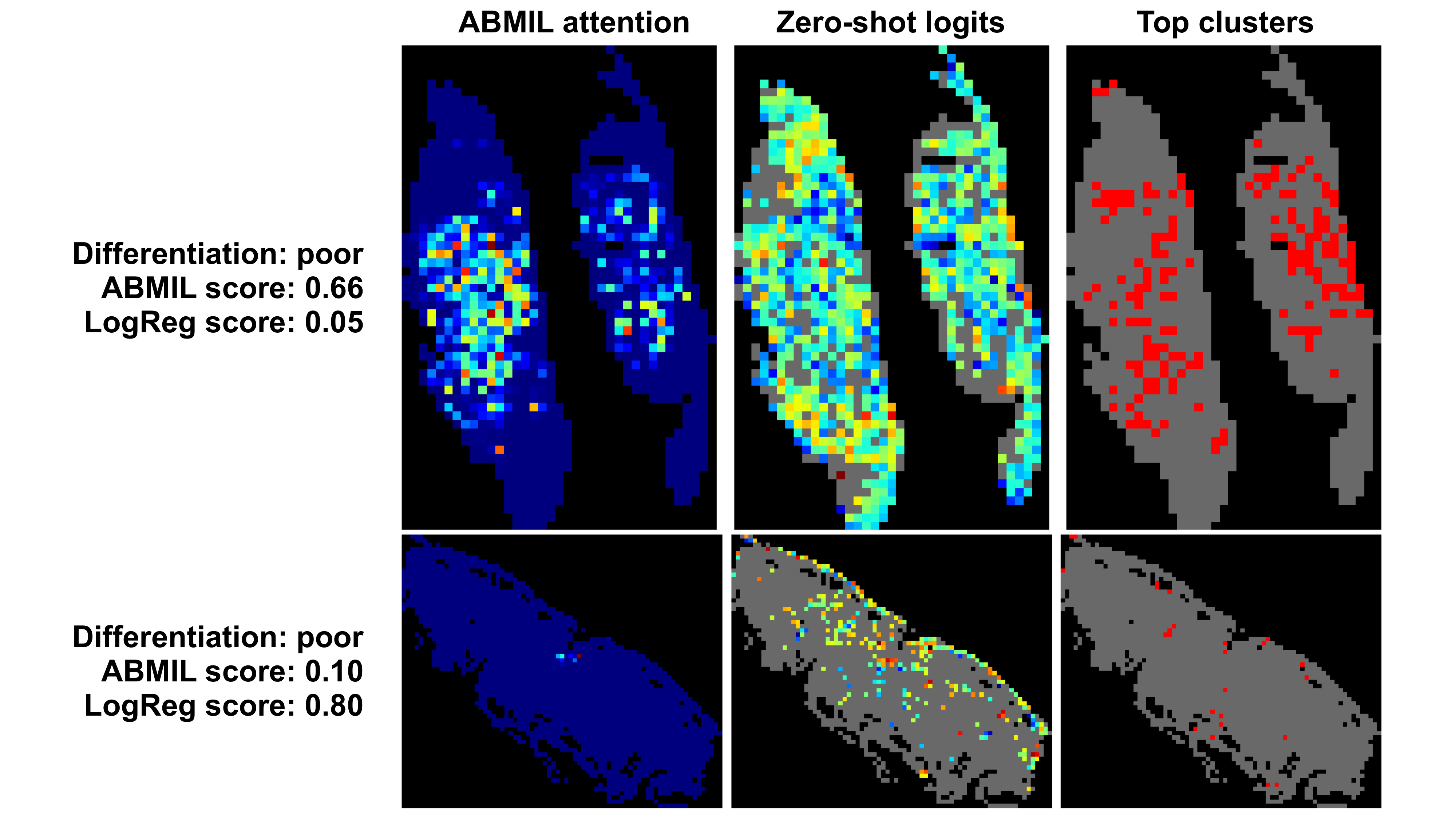}
    \caption{
Comparison of slide-level predictions and spatial patterns across classification methods for two poorly differentiated cSCC tumors. Each row shows one tumor where ABMIL and logistic regression (LogReg) produced divergent scores: in the top row, ABMIL assigned a high score (0.66) while LogReg assigned a low score (0.05); in the bottom row, the reverse pattern was observed (0.10 vs. 0.80). Left: ABMIL attention heatmaps showing normalized attention weights per patch. Middle: Zero-shot logits, defined as the difference in cosine similarity between the phrases ``poorly differentiated'' and ``well differentiated'', shown only for patches classified as cSCC. Right: Patch-level cluster maps highlighting the four clusters most strongly associated with poor differentiation (red) based on univariate AUROC analysis (Methods). Black indicates background, and gray indicates tissue excluded by the corresponding method. All visualizations are based on CONCH embeddings.
    }
    \label{fig:heatmaps}
\end{figure}

Weakly supervised ABMIL models outperformed zero-shot and clustering baselines, reaching AUROCs of 0.81 (95\% CI: 0.71–0.90) and 0.78 (95\% CI: 0.68–0.87) for CONCH and MUSK, respectively (Figure \ref{fig:auroc_logistic_abmil}b, Table \ref{tab:abmil_perf}). Attention heatmaps highlighted regions most influential to the predictions, revealing both overlaps and differences with informative patches identified in the zero-shot and clustering analyses (Figure ~\ref{fig:heatmaps}). Training on finer-grained grade distinctions showed similar trends (Appendix \ref{apd:cscc_grading}).

\subsection{Data efficiency and model generalization}
We further analyzed how model performance scaled with training data. As expected, accuracy improved with larger training sets, with ABMIL models showing greater gains than logistic regression (Figure \ref{fig:auroc_logistic_abmil}). Notably, ABMIL required relatively few training samples ($<$100) to surpass logistic regression performance, particularly when using CONCH embeddings.

\section{Discussion} 
In this manuscript, we introduced \textbf{PathFMTools}, a modular and extensible Python package designed to facilitate the efficient execution, exploration, and visualization of computational pathology foundation models. Using this package, we systematically evaluated two vision-language foundation models, CONCH and MUSK, on the clinically valuable task of histological grading in cutaneous squamous cell carcinoma. Our analysis explored a variety of model adaptation approaches, including zero-shot inference, unsupervised clustering, and weakly-supervised learning, demonstrating clear performance differences and tradeoffs between these methodologies.

We found that zero-shot classification using vision-language embeddings with no task-specific fine-tuning was moderately effective, achieving better-than-chance performance. This highlights the flexibility of vision-language foundation models, especially in contexts where annotated data is limited; however the performance was insufficient for the tested task and formulation. 

Our clustering analysis provided interpretable insights into the morphological patterns learned by foundation models. We demonstrated that individual clusters could be independently informative for predicting histological differentiation, suggesting that foundation models can effectively capture biologically meaningful features. Notably, several identified clusters corresponded to histological features that are known to be relevant to cSCC grading, such as clusters enriched for keratin fragments and poorly differentiated tumor cells. Conversely, some clusters captured imaging artifacts or less relevant features (e.g., sweat glands), underscoring the importance of quality control in computational pathology workflows and the risk of data-driven confounders.

Attention-based multiple instance learning (ABMIL) emerged as the most effective approach for leveraging foundation model embeddings, outperforming both zero-shot and clustering-based logistic regression methods even with minimal training data. Importantly, ABMIL attention heatmaps also provide interpretability at the patch level, which can be interactively visualized with \textbf{PathFMTools}.

Across all analyses, both CONCH and MUSK followed similar performance trends: moderate zero-shot accuracy, improved performance with clustering, and the strongest results with ABMIL. While the absolute performance was comparable between models, CONCH consistently outperformed MUSK at each training set size in the ABMIL setting.

A critical consideration in evaluating foundation models in downstream tasks is what constitutes ``good'' performance. While ABMIL demonstrated the highest overall performance, it is unclear whether the observed levels ($\sim$0.80 AUROC) are sufficient for clinical utility. The high inter-rater variability of cSCC grading may suggest that this level is near the inter-reader noise ceiling, but future studies directly comparing human graders and AI in the same setting would be necessary. Beyond AI standalone performance, a fundamental consideration is how AI assistance affects clinician performance and how AI fits into the clinical workflow~\citep{McNamara2024-oc}. Clinical AI deployment involves many steps, ranging from regulatory considerations and UI engineering, thus our analysis and software package should be viewed as informing research and validation, rather than actual clinical use.

\paragraph{Limitations}
Our study has several limitations. Because the analysis relies on data from a single clinical system, the observed performance may not generalize to other sites or populations. We did not explicitly evaluate sensitivity to staining differences and used each model’s default preprocessing; however, our software allows users to easily define and test custom preprocessing pipelines. Accordingly, the results should be viewed primarily as a demonstration of how our tools can help researchers explore and adapt foundation models for specific clinically relevant tasks. Finally, the software is intended for technically oriented researchers and practitioners, rather than as a no-code or GUI-based tool for users with limited programming experience.


\paragraph{Conclusions}
As foundation models continue to improve and use cases expand, our findings illustrate how \textbf{PathFMTools} streamlines exploration and adaptation of these models, highlighting their promise in addressing critical challenges in histopathology, such as the subjective nature of tumor grading. Our quantitative findings can also help inform the optimal adaption strategies to other clinically-important tasks.

\acks{This research is supported by National Institute of Biomedical Imaging and Bioengineering award R21EB035247 and the National Library of Medicine award R01LM014775.}

\bibliography{main}

@ARTICLE{Jambusaria-Pahlajani2025-vh,
  title     = "{riSCC}: A personalized risk model for the development of poor
               outcomes in cutaneous squamous cell carcinoma",
  author    = "Jambusaria-Pahlajani, Anokhi and Jeanselme, Vincent and Wang,
               David M and Ran, Nina A and Granger, Emily E and Cañueto, Javier
               and Brodland, David G and Carr, David R and Carter, Joi B and
               Carucci, John A and Hirotsu, Kelsey E and Karn, Emily E and
               Koyfman, Shlomo A and Mangold, Aaron R and Muradás Girardi, Fabio
               and Shahwan, Kathryn T and Srivastava, Divya and Vidimos, Allison
               T and Willenbrink, Tyler J and Wysong, Ashley and Lotter, William
               and Ruiz, Emily S",
  journal   = "J. Am. Acad. Dermatol.",
  publisher = "Elsevier BV",
  month     =  feb,
  year      =  2025,
  language  = "en"
}

@InProceedings{clip,
  title = 	 {Learning Transferable Visual Models From Natural Language Supervision},
  author =       {Radford, Alec and Kim, Jong Wook and Hallacy, Chris and Ramesh, Aditya and Goh, Gabriel and Agarwal, Sandhini and Sastry, Girish and Askell, Amanda and Mishkin, Pamela and Clark, Jack and Krueger, Gretchen and Sutskever, Ilya},
  booktitle = 	 {Proceedings of the 38th International Conference on Machine Learning},
  pages = 	 {8748--8763},
  year = 	 {2021},
  editor = 	 {Meila, Marina and Zhang, Tong},
  volume = 	 {139},
  series = 	 {Proceedings of Machine Learning Research},
  month = 	 {18--24 Jul},
  publisher =    {PMLR},
  url = 	 {https://proceedings.mlr.press/v139/radford21a.html}
}

@ARTICLE{Xiang2025-ux,
  title     = "A vision–language foundation model for precision oncology",
  author    = "Xiang, Jinxi and Wang, Xiyue and Zhang, Xiaoming and Xi, Yinghua
               and Eweje, Feyisope and Chen, Yijiang and Li, Yuchen and
               Bergstrom, Colin and Gopaulchan, Matthew and Kim, Ted and Yu,
               Kun-Hsing and Willens, Sierra and Olguin, Francesca Maria and
               Nirschl, Jeffrey J and Neal, Joel and Diehn, Maximilian and Yang,
               Sen and Li, Ruijiang",
  journal   = "Nature",
  publisher = "Springer Science and Business Media LLC",
  pages     = "1--10",
  month     =  jan,
  year      =  2025,
  language  = "en"
}

@ARTICLE{McNamara2024-oc,
  title    = "The clinician-{AI} interface: intended use and explainability in
              {FDA}-cleared {AI} devices for medical image interpretation",
  author   = "McNamara, Stephanie L and Yi, Paul H and Lotter, William",
  journal  = "NPJ Digit Med",
  volume   =  7,
  number   =  1,
  pages    =  80,
  month    =  mar,
  year     =  2024,
  language = "en"
}

@article{
    Pocock2022,
    author = {Pocock, Johnathan and Graham, Simon and Vu, Quoc Dang and Jahanifar, Mostafa and Deshpande, Srijay and Hadjigeorghiou, Giorgos and Shephard, Adam and Bashir, Raja Muhammad Saad and Bilal, Mohsin and Lu, Wenqi and Epstein, David and Minhas, Fayyaz and Rajpoot, Nasir M and Raza, Shan E Ahmed},
    doi = {10.1038/s43856-022-00186-5},
    issn = {2730-664X},
    journal = {Communications Medicine},
    month = {sep},
    number = {1},
    pages = {120},
    publisher = {Springer US},
    title = {{TIAToolbox as an end-to-end library for advanced tissue image analytics}},
    url = {https://www.nature.com/articles/s43856-022-00186-5},
    volume = {2},
    year = {2022}
}

@article{pathml,
    author = {Rosenthal, Jacob and Carelli, Ryan and Omar, Mohamed and Brundage, David and Halbert, Ella and Nyman, Jackson and Hari, Surya N. and Van Allen, Eliezer M. and Marchionni, Luigi and Umeton, Renato and Loda, Massimo},
    title = {Building Tools for Machine Learning and Artificial Intelligence in Cancer Research: Best Practices and a Case Study with the PathML Toolkit for Computational Pathology},
    journal = {Molecular Cancer Research},
    volume = {20},
    number = {2},
    pages = {202-206},
    year = {2022},
    month = {02},
    issn = {1541-7786},
    doi = {10.1158/1541-7786.MCR-21-0665},
    url = {https://doi.org/10.1158/1541-7786.MCR-21-0665},
    eprint = {https://aacrjournals.org/mcr/article-pdf/20/2/202/3391365/202.pdf},
}

@article{zhang2025standardizing,
  title={Accelerating Data Processing and Benchmarking of AI Models for Pathology},
  author={Zhang, Andrew and Jaume, Guillaume and Vaidya, Anurag and Ding, Tong and Mahmood, Faisal},
  journal={arXiv preprint arXiv:2502.06750},
  year={2025}
}

@Article{ElNahhas2024,
  author={El Nahhas, Omar S. M. and van Treeck, Marko and W{\"o}lflein, Georg and Unger, Michaela and Ligero, Marta and Lenz, Tim and Wagner, Sophia J. and Hewitt, Katherine J. and Khader, Firas and Foersch, Sebastian and Truhn, Daniel and Kather, Jakob Nikolas},
  title={From whole-slide image to biomarker prediction: end-to-end weakly supervised deep learning in computational pathology},
  journal={Nature Protocols},
  year={2024},
  month={Sep},
  day={16},
  issn={1750-2799},
  doi={10.1038/s41596-024-01047-2},
  url={https://doi.org/10.1038/s41596-024-01047-2}
}

@ARTICLE{Kingma2014-ad,
  title         = "Adam: A Method for Stochastic Optimization",
  author        = "Kingma, Diederik P and Ba, Jimmy",
  journal       = "arXiv [cs.LG]",
  month         =  dec,
  year          =  2014,
  archivePrefix = "arXiv",
  primaryClass  = "cs.LG"
}

@ARTICLE{Ilse2018-ph,
  title         = "Attention-based Deep Multiple Instance Learning",
  author        = "Ilse, Maximilian and Tomczak, Jakub M and Welling, Max",
  journal       = "arXiv [cs.LG]",
  month         =  feb,
  year          =  2018,
  archivePrefix = "arXiv",
  primaryClass  = "cs.LG"
}

@ARTICLE{otsu,
  author={Otsu, Nobuyuki},
  journal={IEEE Transactions on Systems, Man, and Cybernetics}, 
  title={A Threshold Selection Method from Gray-Level Histograms}, 
  year={1979},
  volume={9},
  number={1},
  pages={62-66},
  doi={10.1109/TSMC.1979.4310076}}

@misc{dinov2,
      title={DINOv2: Learning Robust Visual Features without Supervision}, 
      author={Maxime Oquab and Timothée Darcet and Théo Moutakanni and Huy Vo and Marc Szafraniec and Vasil Khalidov and Pierre Fernandez and Daniel Haziza and Francisco Massa and Alaaeldin El-Nouby and Mahmoud Assran and Nicolas Ballas and Wojciech Galuba and Russell Howes and Po-Yao Huang and Shang-Wen Li and Ishan Misra and Michael Rabbat and Vasu Sharma and Gabriel Synnaeve and Hu Xu and Hervé Jegou and Julien Mairal and Patrick Labatut and Armand Joulin and Piotr Bojanowski},
      year={2024},
      eprint={2304.07193},
      archivePrefix={arXiv},
      primaryClass={cs.CV},
      url={https://arxiv.org/abs/2304.07193}, 
}

@ARTICLE{Choudhary2023-rt,
  title         = "{RACR}-{MIL}: Weakly Supervised Skin Cancer Grading using
                   Rank-Aware Contextual Reasoning on Whole Slide Images",
  author        = "Choudhary, Anirudh and Hwang, Angelina and Kechter, Jacob and
                   Saboo, Krishnakant and Bordeaux, Blake and Bhullar, Puneet
                   and Comfere, Nneka and DiCaudo, David and Nelson, Steven and
                   Johnson, Emma and Swanson, Leah and Murphree, Dennis and
                   Mangold, Aaron and Iyer, Ravishankar K",
  journal       = "arXiv [cs.CV]",
  month         =  aug,
  year          =  2023,
  archivePrefix = "arXiv",
  primaryClass  = "cs.CV"
}

@article{lu2024avisionlanguage,
  title={A visual-language foundation model for computational pathology},
  author={Lu, Ming Y and Chen, Bowen and Williamson, Drew FK and Chen, Richard J and Liang, Ivy and Ding, Tong and Jaume, Guillaume and Odintsov, Igor and Le, Long Phi and Gerber, Georg and others},
  journal={Nature Medicine},
  pages={863–874},
  volume={30},
  year={2024},
  publisher={Nature Publishing Group}
}

@ARTICLE{Schmults2013-qi,
  title    = "Factors predictive of recurrence and death from cutaneous squamous
              cell carcinoma: a 10-year, single-institution cohort study",
  author   = "Schmults, Chrysalyne D and Karia, Pritesh S and Carter, Joi B and
              Han, Jiali and Qureshi, Abrar A",
  journal  = "JAMA Dermatol.",
  volume   =  149,
  number   =  5,
  pages    = "541--547",
  month    =  may,
  year     =  2013,
  language = "en"
}

@ARTICLE{Thompson2016-oo,
  title    = "Risk Factors for Cutaneous Squamous Cell Carcinoma Recurrence,
              Metastasis, and Disease-Specific Death: A Systematic Review and
              Meta-analysis",
  author   = "Thompson, Agnieszka K and Kelley, Benjamin F and Prokop, Larry J
              and Murad, M Hassan and Baum, Christian L",
  journal  = "JAMA Dermatol.",
  volume   =  152,
  number   =  4,
  pages    = "419--428",
  month    =  apr,
  year     =  2016,
  language = "en"
}

@ARTICLE{Que2018-yq,
  title    = "Cutaneous squamous cell carcinoma: Incidence, risk factors,
              diagnosis, and staging",
  author   = "Que, Syril Keena T and Zwald, Fiona O and Schmults, Chrysalyne D",
  journal  = "J. Am. Acad. Dermatol.",
  volume   =  78,
  number   =  2,
  pages    = "237--247",
  month    =  feb,
  year     =  2018,
  language = "en"
}

@ARTICLE{Nash2022-dd,
  title    = "Grading of differentiation in cutaneous squamous cell carcinoma:
              Evaluation of interrater and intrarater reliability",
  author   = "Nash, Jessica and Shahwan, Kathryn T and Chung, Catherine and
              Abidi, Nadia and Gokun, Yevgeniya and Pan, Xueliang and Carr,
              David R",
  journal  = "J. Am. Acad. Dermatol.",
  volume   =  87,
  number   =  4,
  pages    = "895--897",
  month    =  oct,
  year     =  2022,
  language = "en"
}

@misc{li2025review,
      title={A Survey on Computational Pathology Foundation Models: Datasets, Adaptation Strategies, and Evaluation Tasks}, 
      author={Dong Li and Guihong Wan and Xintao Wu and Xinyu Wu and Ajit J. Nirmal and Christine G. Lian and Peter K. Sorger and Yevgeniy R. Semenov and Chen Zhao},
      year={2025},
      eprint={2501.15724},
      archivePrefix={arXiv},
      primaryClass={cs.CV},
      url={https://arxiv.org/abs/2501.15724}, 
}

@misc{bilal2025review,
      title={Foundation Models in Computational Pathology: A Review of Challenges, Opportunities, and Impact}, 
      author={Mohsin Bilal and Aadam and Manahil Raza and Youssef Altherwy and Anas Alsuhaibani and Abdulrahman Abduljabbar and Fahdah Almarshad and Paul Golding and Nasir Rajpoot},
      year={2025},
      eprint={2502.08333},
      archivePrefix={arXiv},
      primaryClass={cs.CV},
      url={https://arxiv.org/abs/2502.08333}, 
}

\clearpage
\appendix

\section{}\label{apd:first}
\subsection{Cohort Summary}\label{apd:cohort}
\begin{table}[h]
\centering 
\small
  \caption{Cohort Summary. IQR: interquartile range.}
\begin{tabular}{lll}
\textbf{}                  &                    & \textbf{Value (\%)} \\
\hline
\textbf{Total}               &   Tumor count     & 440    \\
&   Patient count     & 412    \\
\hline
\textbf{Patient Sex}       & Male               & 280 (64)             \\
                           & Female             & 160 (36)             \\
\hline
\textbf{Age}               & Median (IQR)       & 74 (66, 82)    \\
\hline
\textbf{Tumor Location}    & Head/neck          & 215 (49)             \\
                           & Trunk/extremeties  & 224 (51)             \\
                           & Unknown            & 1 (\textless{}1)     \\
\hline
\textbf{Diameter}          & Median (IQR), mm   & 13 (8, 23)           \\
\hline
\textbf{Invasion Depth} & Dermis             & 217 (49)             \\
                           & Subcutaneous fat   & 47 (11)              \\
                           & Beyond subcut. fat & 41 (9)               \\
                           & Unknown            & 135 (31)             \\
\hline
\textbf{Differentiation}   & Well               & 239 (54)             \\
                           & Moderate           & 99 (23)              \\
                           & Poor               & 102 (23)            
\end{tabular}
\label{cohort_counts}
\end{table}
\subsection{ABMIL approach}\label{apd:abmil_math}
 Given a set of $n$ $d$-dimensional patch embeddings $\{z_1,\ \dots,\ z_n\} \in \mathbb{R}^d$ for a given slide, the model computes attention weights $\alpha_i \in [0, 1]$ as $$\alpha_i = \frac{\exp \bigg( w^T \tanh(Vz_i^T) \odot \sigma(Uz_i^T)\bigg)}{\sum_{j=1}^n \exp \bigg( w^T \tanh(Vz_j^T) \odot \sigma(Uz_j^T) \bigg)}$$
where $\sigma$ is the sigmoid function, $V, U \in \mathbb{R}^{h \times d}$ and $w \in \mathbb{R}^h$ are learned parameters, $h=256$ is the attention hidden dimensionality, and $d$ is the dimension of the embedding produced by the foundation model. 

The slide-level representation is then computed as the attention-weighted sum of patch embeddings $$z_{slide}:= \sum_{i=1}^n \alpha_iz_i \in \mathbb{R}^d$$

\subsection{Hyperparameters \& cross validation details}\label{apd:abmil_hyper_cross}
We trained the ABMIL models using the Adam optimizer~\citep{Kingma2014-ad} and a cyclic learning rate that varied linearly from $5 \times 10^{-5}$ to $5 \times 10^{-4}$, using a half-cycle length of 2 epochs. Logistic regression models were trained using $\ell_2$ regularization with a penalty scaling factor of 1 and a maximum of 1,000 iterations.

To evaluate model performance, we performed 5-fold cross-validation at the patient level for both the ABMIL and logistic regression models. In each fold, 20\% of the data was held out as a test set, with stratified sampling to preserve the proportion of positive and negative slides across splits and prevent different tumors from a single patient from appearing in both the development and test splits of a given fold. For ABMIL, an additional 20\% of the data was held out as a validation set for monitoring model convergence.

We computed AUROC scores on the test set within each fold, and reported the mean across folds as the primary performance estimate. To quantify variability, we computed 95\% DeLong confidence intervals for the AUROC point estimates. Furthermore, to assess data efficiency, we evaluated model performance as a function of training set size. For both ABMIL and logistic regression, we trained models on progressively larger subsets of the training data, using increments of $10\%$ of the full dataset.

\subsection{cSCC grading performance}\label{apd:cscc_grading}
When considering individual cSCC grades, the ABMIL models achieved the highest accuracy when distinguishing well- from poorly differentiated tumors, whereas classification between moderately and poorly differentiated tumors was more challenging but still yielded reasonable performance. Table~\ref{tab:abmil_perf} summarizes AUROC values across different grading cutoffs.

\begin{table}[h]
    \centering
    \small
    \begin{tabular}{lllcc}
        \hline
        Positive class & Negative class & & \multicolumn{2}{c}{AUROC} \\
        \cline{4-5}
        & & & CONCH & MUSK \\
        \hline
        Moderate & Well & & \textbf{0.74} & 0.71 \\
        Poor & Well & & \textbf{0.87} & 0.82 \\
        Poor & Moderate & & \textbf{0.73} & 0.66 \\
        Moderate \& Poor & Well & & \textbf{0.81} & 0.78 \\
        \hline
    \end{tabular}
    \caption{AUROC of ABMIL models using CONCH and MUSK embeddings for binary cSCC grading using different groups. The higher AUROC for each comparison is shown in bold.}
    \label{tab:abmil_perf}
\end{table}

Table \ref{tab:sens_spec_grouped} summarizes the sensitivity and specificity of the CONCH and MUSK models at three operating points: high-sensitivity (targeting 90\%), high-specificity (targeting 90\%), and balanced (targeting equal sensitivity and specificity). For each fold, thresholds were calibrated on the validation set and then applied to the corresponding test set. Point estimates and 95\% confidence intervals (Adjusted Wald method) were obtained by aggregating predictions across all test folds.

\begin{table}[h]
\centering
\small
\resizebox{\columnwidth}{!}{%
\begin{tabular}{lrrr}
\hline
\textbf{Operating Point} & \textbf{} & \textbf{CONCH} & \textbf{MUSK} \\
\hline
\textbf{High Sensitivity} & & & \\
\hspace{1em}\textbf{Sens} & & 0.91 (0.89–0.92) & 0.84 (0.83–0.85) \\
\hspace{1em}\textbf{Spec} & & 0.45 (0.43–0.46) & 0.50 (0.48–0.52) \\
\hline
\textbf{High Specificity} & & & \\
\hspace{1em}\textbf{Sens} & & 0.59 (0.57–0.61) & 0.46 (0.44–0.48) \\
\hspace{1em}\textbf{Spec} & & 0.89 (0.88–0.90) & 0.84 (0.83–0.86) \\
\hline
\textbf{Balanced} & & & \\
\hspace{1em}\textbf{Sens} & & 0.74 (0.72–0.76) & 0.74 (0.72–0.76) \\
\hspace{1em}\textbf{Spec} & & 0.74 (0.72–0.75) & 0.64 (0.62–0.66) \\
\hline
\end{tabular}%
}
\caption{Sensitivity and specificity for CONCH and MUSK models at three operating points. Thresholds were calibrated per fold on validation data and evaluated on held-out test folds.}
\label{tab:sens_spec_grouped}
\end{table}

\begin{figure*}
    \centering
    \includegraphics[width=1\linewidth]{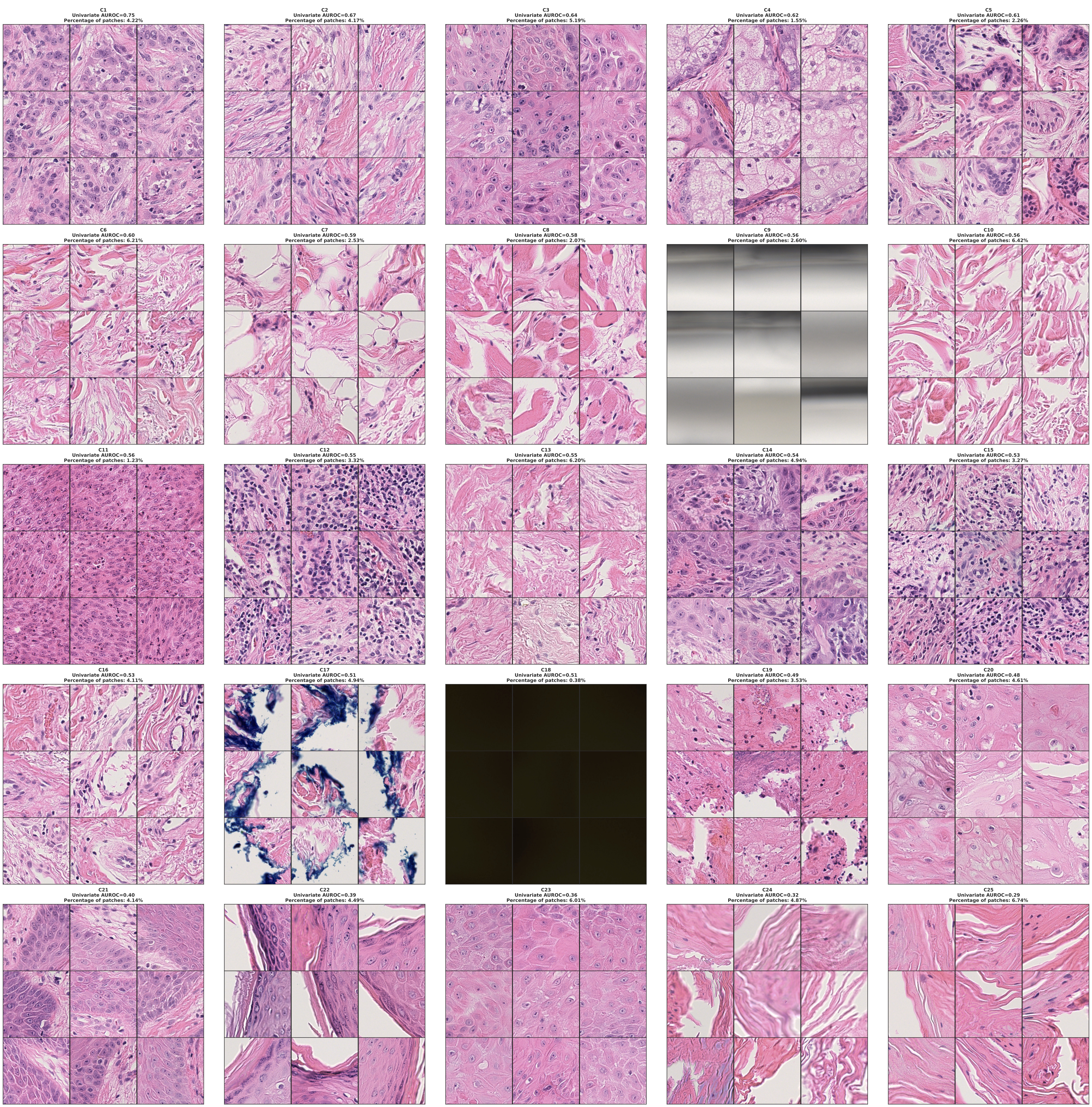}
    \caption{
    Representative patches from each of the 25 CONCH embedding clusters. The outer grid represents individual clusters, each containing an inner $3 \times 3$ grid of example patches nearest to the cluster centroid. Clusters are ordered left to right, top to bottom by decreasing univariate AUROC for predicting cancer grade (Methods), with AUROC values indicated above each cluster. The text annotation also specifies the percentage of total patches assigned to each cluster. 
    }
    \label{fig:conch_clusters_full}
\end{figure*}

\begin{figure*}
    \centering
    \includegraphics[width=1\linewidth]{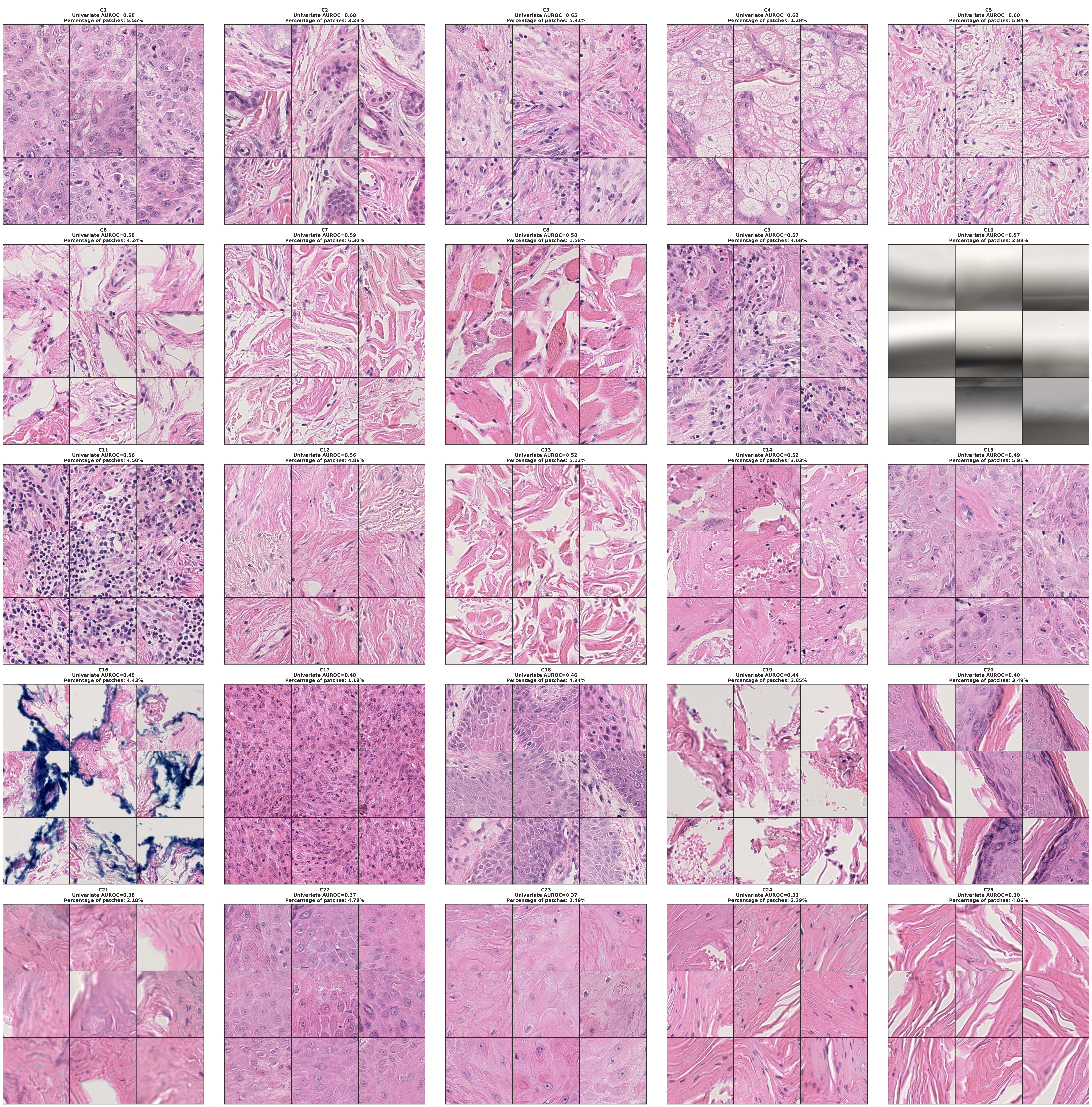}
    \caption{
       Representative patches from each of the 25 MUSK embedding clusters. The outer grid represents individual clusters, each containing an inner $3 \times 3$ grid of example patches nearest to the cluster centroid. Clusters are ordered left to right, top to bottom by decreasing univariate AUROC for predicting cancer grade (Methods), with AUROC values indicated above each cluster. The text annotation also specifies the percentage of total patches assigned to each cluster. 
    }
    \label{fig:musk_clusters_full}
\end{figure*}

\subsection{Runtime analysis}\label{apd:runtime}
All experiments were conducted on a server equipped with an NVIDIA H100 GPU, an Intel Xeon Silver 4410Y CPU, running Linux, with data stored on a network file system (NFS). Table \ref{tab:profiling} summarizes the average time per slide for each major processing stage, measured across all slides in the cohort.

\begin{table}[h]
    \centering
    \small
    \begin{tabular}{l c}
        \hline
        \textbf{Step} & \textbf{Per slide (mean $\pm$ SD)} \\
        \hline
        Segmentation     & 4.2 $\pm$ 3.8 s \\
        Tiling           & 20.1 $\pm$ 25.7 s \\
        CONCH embedding  & 23.2 $\pm$ 25.5 s \\
        MUSK embedding   & 46.9 $\pm$ 57.0 s \\
        \hline
    \end{tabular}
    \caption{Per-slide runtime (mean $\pm$ SD) for each stage of the slide processing pipeline. Times were averaged across all slides.}
    \label{tab:profiling}
\end{table}

\subsection{Sensitivity to prompt choice for zero-shot analysis}\label{apd:prompt}
The zero-shot results presented in the main text followed a clinically motivated strategy where we first identified patches with a higher logit for ``cutaneous squamous cell carcinoma'' than for ``non-neoplastic'', and then computed a slide-level score as the maximum logit difference between ``poorly differentiated'' and ``well differentiated'' across these patches. Here, we also present results using alternative aggregation methods rather than the maximum. Specifically, we (1) averaged the logit differences across all identified patches and (2) averaged them across the 10\% of identified patches with the largest logit difference. Both methods yielded results comparable to the clinically motivated approach, with CONCH achieving AUROCs of 0.66 and 0.64, and MUSK achieving 0.58 and 0.61, respectively.  

We also tested prompt ensembling, averaging logits over ten semantically varied prompt phrasings for each class (Table \ref{tab:prompt_variants}). Ensembling did not improve performance (AUROC $\sim$0.55 for both CONCH and MUSK), potentially reflecting the consistent terminology used in cSCC diagnosis.

\begin{table}[h]
\centering
\small
\begin{tabular}{p{0.25\linewidth} p{0.68\linewidth}}
\toprule
\textbf{Label} & \textbf{Prompt variants} \\
\midrule
Cutaneous squamous cell carcinoma &
cutaneous squamous cell carcinoma; cSCC; cutaneous SCC; squamous cell carcinoma of the skin;
squamous carcinoma; cutaneous squamous carcinoma; H\&E image of cutaneous squamous cell carcinoma;
histopathology image of skin squamous cell carcinoma; tumor region consistent with cSCC;
slide patch showing squamous carcinoma. \\[3pt]

Non-neoplastic &
non-neoplastic; benign tissue; normal skin; non-tumorous region; non-cancerous tissue;
normal cells; H\&E image of normal skin; tissue without neoplastic changes;
non-lesional tissue patch; normal tissue. \\[3pt]

Well differentiated &
well differentiated; well-differentiated carcinoma; well-differentiated squamous cell carcinoma;
well-organized tumor architecture; low-grade carcinoma; low grade; well differentiated tumor cells;
tumor with minimal atypia; slide showing well-differentiated tumor cells; well-differentiated tumor. \\[3pt]

Poorly differentiated &
poorly differentiated; poorly-differentiated carcinoma; high-grade squamous cell carcinoma;
undifferentiated tumor cells; disorganized tumor architecture; marked cellular atypia;
aggressive poorly differentiated SCC; slide showing poorly differentiated cells; high grade;
poorly differentiated tumor cells. \\
\bottomrule
\end{tabular}
\caption{Prompt variants used for each class.}
\label{tab:prompt_variants}
\end{table}

\subsection{Example code}\label{apd:Code}
With PathFMTools, common computational pathology workflows can be performed using a few lines of code, as highlighted in the following example:
\begin{lstlisting}
# Instantiate one Slide object per WSI
slides = [Slide(fp) for fp in PATHS]
for slide in slides:
    # Apply preprocessing (segmentation + tiling)
    slide.preprocess(PATCH_SIZE, SEG_METHOD)
    # Run foundation model inference on slide patches
    slide.embed_patches(MODEL)
    # Run zero-shot classification on slide patches
    slide.classify_patches(MODEL, PROMPTS)
# Perform K-means clustering on all patches from
# all WSIs jointly
slide_group = SlideGroup(slides)
kmeans = KMeansPatchClusterer.fit_predict(slide_group, k=K)
\end{lstlisting}

\end{document}